# *HeadOn*: Real-time Reenactment of Human Portrait Videos

JUSTUS THIES, Technical University of Munich
MICHAEL ZOLLHÖFER, Stanford University
CHRISTIAN THEOBALT, Max-Planck-Institute for Informatics
MARC STAMMINGER, University of Erlangen-Nuremberg
MATTHIAS NIESSNER, Technical University of Munich

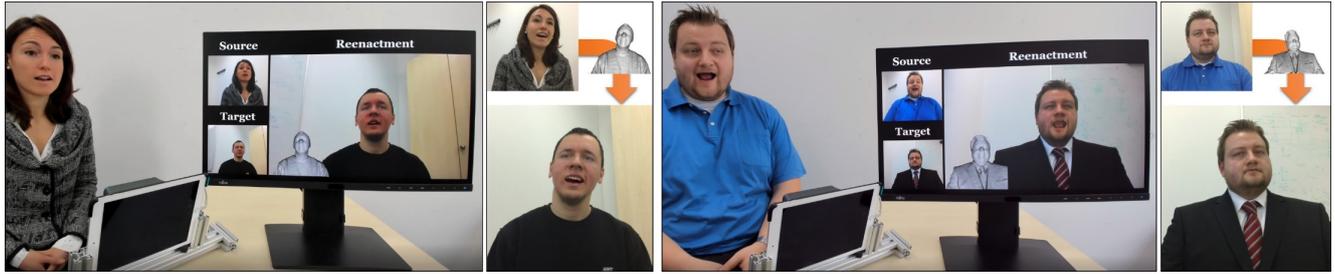

Fig. 1. Our novel *HeadOn* approach enables real-time reenactment of upper body motion, head pose, face expression, and eye gaze in human portrait videos. For synthesis of new photo-realistic video content, we employ a novel video-based rendering approach that builds on top of a fully controllable 3D actor model. The person-specific model is constructed from a short RGB-D calibration sequence and is driven by a real-time torso and face tracker.

We propose HeadOn, the first real-time source-to-target reenactment approach for complete human portrait videos that enables transfer of torso and head motion, face expression, and eye gaze. Given a short RGB-D video of the target actor, we automatically construct a personalized geometry proxy that embeds a parametric head, eye, and kinematic torso model. A novel real-time reenactment algorithm employs this proxy to photo-realistically map the captured motion from the source actor to the target actor. On top of the coarse geometric proxy, we propose a video-based rendering technique that composites the modified target portrait video via view- and pose-dependent texturing, and creates photo-realistic imagery of the target actor under novel torso and head poses, facial expressions, and gaze directions. To this end, we propose a robust tracking of the face and torso of the source actor. We extensively evaluate our approach and show significant improvements in enabling much greater flexibility in creating realistic reenacted output videos.

Authors' addresses: Justus Thies, Technical University of Munich, justus.thies@tum.de; Michael Zollhöfer, Stanford University, zollhoefer@cs.stanford.edu; Christian Theobalt, Max-Planck-Institute for Informatics, theobalt@mpi-inf.mpg.de; Marc Stamminger, University of Erlangen-Nuremberg, marc.stamminger@fau.de; Matthias Nießner, Technical University of Munich, niessner@tum.de.

## 1 INTRODUCTION

Reenactment approaches aim to transfer the motion of a source actor to an image or video of a target actor. Very recently, facial reenactment methods have been successfully employed to achieve highly-realistic manipulations of facial expressions based on commodity video data [Averbuch-Elor et al. 2017; Suwajanakorn et al. 2017; Thies et al. 2015, 2016, 2018; Vlasic et al. 2005]. Rather than animating a virtual, stylized avatar (e.g., as used in video games), these algorithms replace the face region of a person with a synthetic re-rendering, or modify the target image/video under guidance of a 3D face model. This enables changing the expression of a target person and creating a manipulated output video that suggests different content; e.g., a person who is sitting still could appear as if he/she is talking. Modern reenactment approaches achieve highly believable results, even in real-time, and have enjoyed wide media coverage due to the interest in general movie and video editing [Vlasic et al. 2005], teleconferencing [Thies et al. 2018], reactive profile pictures [Averbuch-Elor et al. 2017], or visual dubbing of foreign language movies [Garrido et al. 2015].

Even though current facial reenactment results are impressive, they are still fundamentally limited in the type of manipulations they enable. For instance, these approaches are only able to modify facial expressions, whereas the rigid pose of the head, including its orientation, remains unchanged and does not follow the input video. Thus, only subtle changes, such as opening the mouth or adding wrinkles on the forehead are realized, which severely limits the applicability to video editing, where the control of the pose of the target person is also required. Furthermore, without joint modification of the head pose, the modified facial expressions often seem out-of-place, since they do not well align with visual pauses in the body and head motion; as noted by Suwajanakorn et al. [2017] this significantly restricts the applicability in teleconferencing scenarios.

In this work, we thus go one step further by introducing *HeadOn*, a reenactment system for portrait videos recorded with a commodity RGB-D camera. We overcome the limitations of current facial reenactment methods by not only controlling changes in facial expression, but also reenacting the rigid position of the head, of the upper body, and the eye gaze – i.e., the entire person-related content in a portrait video.

At the core of our approach is the combination of robust and accurate tracking of a deformation proxy with view-dependent texturing for video-based re-rendering. To achieve this, we propose a new method to swiftly and automatically construct a personalized head and torso geometry proxy of a human from a brief RGB-D initialization sequence. The shape proxy features a personalized parametric 3D model of the complete head that is rigged with blendshapes for expression control and is integrated with a personalized upper torso model. A new real-time reenactment algorithm employs this proxy to photo-realistically map face expression and eye gaze, as well as head and torso motion of a captured source actor to a target actor. To this end, we contribute a new photo-realistic video-based rendering approach that composites the reenacted target portrait video via view- and pose-dependent texturing and video compositing.

In summary, we contribute the following:

- rapid automatic construction of a personalized geometry proxy that embeds a parametric human face, eye, full head, and upper body model,
- a photo-realistic, view-, and pose-dependent texturing and compositing approach,
- a robust tracking approach of the source actor,
- and real-time source-to-target reenactment of complete human portrait videos.

## 2 RELATED WORK

Face reconstruction and reenactment have a long history in computer graphics and vision. We focus on recent approaches based on lightweight commodity sensors. For an overview of high-quality techniques that use controlled acquisition setups, we refer to Klehm et al. [2015]. Recently, a state-of-the-art report on monocular 3d face reconstruction, tracking and applications has been published that gives a comprehensive overview of current methods [Zollhöfer et al. 2018]. In the following we concentrate on the most related techniques.

*Parametric Face Representations.* Current state-of-the-art monocular face tracking and reconstruction approaches heavily rely on 3D parametric identity [Blanz et al. 2003; Blanz and Vetter 1999] and expression models [Tena et al. 2011] that generalize active appearance models [Cootes et al. 2001] from 2D to 3D space. Even combinations of the two have been proposed [Xiao et al. 2004]. Recently, large-scale models in terms of geometry [Booth et al. 2016] and texture [Zafeiriou et al. 2017] have been constructed based on an immense amount of training data (10,000 scans). For modeling facial expressions, the de facto standard in the industry are blendshapes [Lewis et al. 2014; Pighin et al. 1998]. Physics-based models [Ichim et al. 2017; Sifakis et al. 2005] have been proposed in research, but fitting such complex models to commodity video at real-time rates is still challenging. Some approaches [Shi et al. 2014a; Vlasic et al. 2005] jointly represent face identity and expression in a single multi-linear model. Joint shape and motion models [Li et al. 2017] have also been learned from a large collection of 4D scan data. Other approaches [Garrido et al. 2016] reconstruct personalized face rigs, including reflectance and fine-scale detail from monocular video. Liang et al. [Liang et al. 2014] reconstruct the identity of a face from monocular Kinect data using a part-based matching algorithm. They select face parts (eyes,nose,mouth,cheeks) from a database of faces that best match the input data. To get an improved and personalized output they fuse these parts with the Kinect depth data. Ichim et al. [2015] propose to reconstruct 3D avatars from multi-view images recorded by a mobile phone and personalize the expression space using a calibration sequence.

*Commodity Face Reconstruction and Tracking.* The first commodity face reconstruction approaches that employed lightweight capture setups, i.e., stereo [Valgaerts et al. 2012], RGB [Fyffe et al. 2014; Garrido et al. 2013; Shi et al. 2014a; Suwajanakorn et al. 2014, 2015], or RGB-D [Chen et al. 2013] cameras had slow off-line frame rates and required up to several minutes to process a single input frame. These methods either deform a personalized template mesh [Suwajanakorn et al. 2014, 2015; Valgaerts et al. 2012], use a 3D template and expression blendshapes [Fyffe et al. 2014; Garrido et al. 2013], a template and an underlying generic deformation graph [Chen et al. 2013], or additionally solve for the parameters of a multi-linear face model [Shi et al. 2014a]. Suwajanakorn et al. [2014; 2015] build a modifiable mesh model from internet photo collections. Shi et al. [2014b] key-frame based bundle adjustment to fit the multi-linear model. Recently, first methods have appeared that reconstruct facial performances in real-time from a single commodity RGB-D camera [Bouaziz et al. 2013; Hsieh et al. 2015; Li et al. 2013; Thies et al. 2015; Weise et al. 2011; Zollhöfer et al. 2014]. Dense real-time face reconstruction has also been demonstrated based on monocular RGB data using trained regressors [Cao et al. 2014a, 2013] or analysis-by-synthesis [Thies et al. 2016]. Even fine-scale detail can be recovered at real-time frame rates [Cao et al. 2015].

*Performance Driven Facial Animation.* Face tracking has been applied to control virtual avatars in many contexts. First approaches were based on sparse detected feature points [Chai et al. 2003; Chuang and Bregler 2002]. Current methods for character animation [Cao et al. 2015, 2014a, 2013; Weise et al. 2009], teleconferences [Weise et al. 2011], games [Ichim et al. 2015], and virtual reality [Li et al. 2015; Olszewski et al. 2016] are based on dense alignment energies. Olszewski et al. [2016] proposed an approach to control a digital avatar in real-time based on an HMD-mounted RGB camera. Recently, Hu et al. [2017] reconstructed a stylized 3D avatar, including hair, from a single image that can be animated and displayed in virtual environments. General image-based modeling and rendering techniques [Gortler et al. 1996; Isaksen et al. 2000; Kang et al. 2006; Kopf et al. 2013; Wood et al. 2000] enable the creation of photo-realistic imagery for many real-world effects that are hard to render and reconstruct at a sufficiently high quality using current approaches. In the context of portrait videos, especially fine details, e.g., single strands of hair or high-quality apparel, are hard to reconstruct. Cao et al. [2016] drive dynamic image-based 3D avatars based on a real-time face tracker. We go one step further and combine a controllable geometric actor rig with video-based rendering techniques to enable the real-time animation and synthesis of a photo-realistic portrait video of a target actor.

*Face Reenactment and Replacement.* Face reconstruction and tracking enabling the manipulation of faces in videos has already found its way into consumer applications, e.g., Snapchat, Face Changer, and FaceSwap. Face replacement approaches [Dale et al. 2011; Garrido et al. 2014] swap out the facial region of a target actor and replace it with the face of a source actor. Face replacement is also possible in portrait photos crawled from the web [Kemelmacher-Shlizerman 2016]. In contrast, facial reenactment approaches preserve the identity of the target actor and modify only the facial expressions. The first approaches worked offline [Vlasic et al. 2005] and required controlled recording setups. Thies et al. [2015] proposed the first real-time expression mapping approach based on an RGB-D camera. Follow-up works enabled real-time reenactment of monocular videos [Thies et al. 2016] and stereo video content [Thies et al. 2017, 2018]. Visual video dubbing approaches try to match the mouth motion to a dubbed audio-track [Garrido et al. 2015]. For mouth interior synthesis, image-based [Kawai et al. 2014; Thies et al. 2016] and template-based [Thies et al. 2015] approaches have been proposed. Recently, Suwajanakorn et al. [2017] presented an impressive system mapping audio input to plausible lip motion using a learning-based approach. Even though all of these approaches obtain impressive results, they are fundamentally limited in the types of enabled manipulations. For instance, the rigid pose of the upper body and head cannot be modified. One exception is the offline approach of Elor et al. [Averbuch-Elor et al. 2017] that enables the creation of reactive profile videos while allowing mapping of small head motions based on image warping. Our approach goes one step further by enabling complete reenactment of portrait videos, i.e., it enables larger changes of the head pose, control over the torso, facial reenactment and eye gaze redirection, all at real-time frame rates, which is of paramount importance for live teleconferencing scenarios.

Recently, Ma et al. [Ma et al. 2017] proposed a generative framework that allows to synthesize images of people in novel body poses. They employ a U-Net-like generator that is able to synthesize images at a resolution of 256 × 256 pixels. While showing nice results, they only work on single images and not videos; they are not able to modify facial expressions.

## 3 METHOD OVERVIEW

Our approach is a synergy between many tailored components. In this section we give an overview of our approach; before explaining all components in the following sections. Fig. 2 depicts the pipeline of the proposed technique. We distinguish between the source actor and the target actor that has to be reenacted using the expressions and motions of the source actor. The source actor is tracked in real time using a dense face tracker and a model-to-frame Iterative Closest Point (ICP) method to track the torso of the person (details are given in Sec. 6.1). To be able to transfer the expressions and the rigid motion of the head as well as the torso to the target actor, we construct a video-based actor rig (see Sec. 4). This actor rig is based on the combination of the SMPL body model [Loper et al. 2015] and a parametric face model that is also used to track the facial expressions of the source actor. Our novel video-based rendering technique (Sec. 5) allows us to render the target actor rig in a photo-realistic fashion. Since the face model used to rig the target actor is the same as the model used to track the source actor, we can directly copy the expression parameters from the source model to the target rig. To transfer the body pose, we compute the relative pose between the tracked face and the torso. Using inverse kinematic we map the pose to the three involved joints of the SMPL skeleton (head, neck and torso joint; each having three degrees of freedom). In Sec. 7 and in the supplemental video we demonstrate the effectiveness of our technique and we compare our results against state-of-the-art approaches.

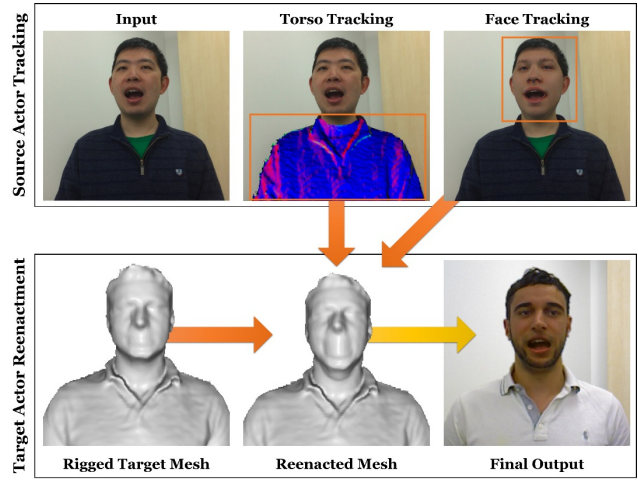

Fig. 2. Overview of our proposed *HeadOn* technique. Based on the tracking of the torso and the face of the source actor, we deform the target actor mesh. Using this deformed proxy of the target actor's body, we use our novel view-dependent texturing to generate a photo-realistic output.

## 4 GENERATING A VIDEO-BASED ACTOR RIG

The first key component of our approach is the fully automatic generation of a video-based person-specific rig of the target actor from commodity RGB-D input. The actor rig combines a unified parametric representation of the target's upper body (chest, shoulders, and neck, no arms) and head geometry with a video-based rendering technique that enables the synthesis of photo-realistic portrait video footage. In this section, we describe the reconstruction of a fully rigged geometric model of the target actor. This model is then used as a proxy for video-based re-rendering of the target actor, as described in Section 5. Fig. 3 shows an overview of the actor rig generation pipeline.

### 4.1 Input Data Acquisition

As input, we record two short video sequences of the target actor. The first sequence is a short stream $S = \{C_t, D_t\}_t$ of color $C_t$ and depth $D_t$ images of the target actor under different viewing angles.

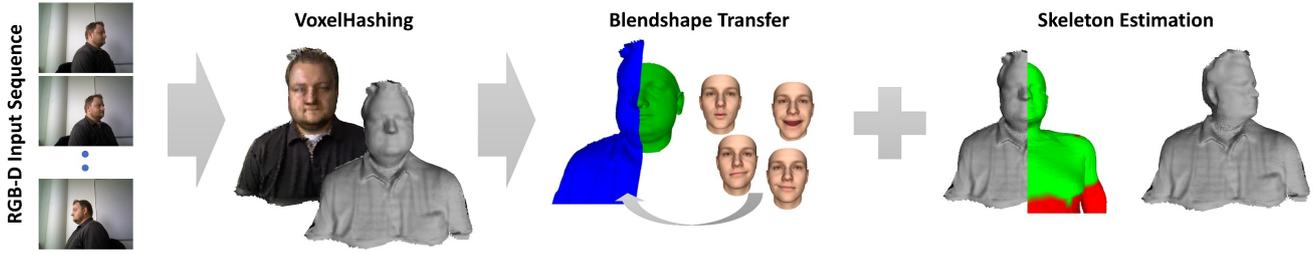

Fig. 3. Automatic generation of a fully controllable person-specific target actor rig. We reconstruct a coarse geometric proxy of the torso and head based on a commodity RGB-D stream. To gain full parametric control of the target actor, we automatically rig the model with facial expression blendshapes and a kinematic skeleton.

We assume that the target actor is sitting on a swivel chair and is initially facing the camera. The target actor then first rotates the chair to a left profile view (−90°), followed by a right profile view (+90°), while keeping the body and head pose as rigid as possible. Our starting pose is camera-facing to enable robust facial landmark detection in the first frame, which is required for later registration steps. Based on this sequence, we generate most parts of our actor rig, except eye gaze control, for which we need an additional recording of the eye motion. In this sequence, the target actor faces the camera and looks at a moving dot on a screen directly in front of him. The actor follows the dot with his eyes without moving the head. This sequence is used for an eye gaze transfer strategy similar to Thies et al. [2017; 2018]. The complete recording of these two datasets takes less than 30 seconds, with approximately 10 seconds for the body and 20 seconds for the eye data acquisition step. Note, we only capture images of the person in a single static pose. In particular, we do not capture neck motions.

### 4.2 Reconstruction of the Upper Body and Head Proxy

We start with the reconstruction of the geometry of the torso and head of the target actor, based on the recorded depth images $D_t$ of the body sequence. First, we estimate the rigid pose of the actor in each frame, relative to the canonical pose in the first frame, using projective data association and an iterative closest point (ICP) [Besl and McKay 1992; Chen and Medioni 1992] strategy based on a point-to-plane distance metric [Low 2004]. We then fuse all depth observations $D_t$ in a canonical truncated signed distance (TSDF) representation [Curless and Levoy 1996; Newcombe et al. 2011]. We are using the open source VoxelHashing [Nießner et al. 2013] implementation that stores the TSDF in a memory efficient manner to reconstruct the actor in its canonical pose. In all our experiments, we use a voxel size of 4 mm. Finally, we extract a mesh using Marching Cubes [Lorensen and Cline 1987].

For every tracked frame, we also store the rigid transformation of the body with respect to the canonical pose, which we need for view-dependent texturing in a later step. For the eye calibration sequence, we also estimate the rigid pose for each frame, by fitting the previously obtained model using a projective point-to-plane ICP. We need these poses later to enable the re-projection of the eyes in the synthesis stage.

### 4.3 Multi-linear Face Model to Scan Fitting

To gain full parametric control of the person-specific actor model, we automatically rig the reconstructed mesh. To this end, we first fit a statistical morphable face model to establish correspondence and then transfer facial blendshapes to the actor model. We use the multi-linear face model of [Thies et al. 2016] that is based on the statistical face model of Blanz and Vetter [Blanz and Vetter 1999] and the blendshapes of [Alexander et al. 2009; Cao et al. 2014b].

*Sparse Feature Alignment.* The used model-based non-rigid registration approach is based on a set of sparse detected facial feature points and a dense geometric alignment term. The sparse discriminative feature points are detected in the frontal view of the body calibration sequence using the True Vision Solution (TVS) feature tracker[1]. This landmark tracker is a commercial implementation of Saragih et al. [2011]. We lift the detected feature points to 3D by projecting them onto the target proxy mesh using the recovered rigid pose and the known camera intrinsics. The corresponding 3D feature points on the template face are selected once in a preprocessing step and stay constant for all experiments. The sparse feature alignment term is defined as:

$$E_{\text{sparse}}(\alpha, \delta, R, t) = \sum_{(i,j) \in C_{\text{sparse}}} \left\| \left[ R v_i(\alpha, \delta) + t \right] - p_j \right\|_2^2 .$$

Here, $\alpha$ is the vector containing the $N_\alpha = 80$ shape coefficients of the face model and the $\delta$ are the $N_\delta = 76$ blendshape expression weights. We include blendshapes during optimization to compensate for non-neutral face expression of the actor. $R$ is the rotation and $t$ the translation of the face model. The $p_j$ are the points on the proxy mesh and the $v_i(\alpha, \delta)$ are the corresponding sparse points on the template mesh that are computed by a linear combination of the shape and expression basis vectors of the underlying face model. The tuples $(i, j) \in C_{\text{sparse}}$ define the set of feature correspondences.

*Dense Point-to-Point Alignment.* In addition to this sparse feature alignment term, we employ a dense point-to-point alignment energy based on closest point correspondences:

$$E_{\text{dense}}(\alpha, \delta, R, t) = \sum_{(i,j) \in C_{\text{dense}}} \left\| \left[ R v_i(\alpha, \delta) + t \right] - p_j \right\|_2^2 .$$

---

[1]http://truevisionsolutions.net/

The closest point correspondences $C_{\text{dense}}$ are computed using the approximate nearest neighbor (ANN) library[2]. We prune correspondences based on a distance threshold ($thres_{\text{dist}} = 10\,\text{cm}$) and on the orientation of the normals.

*Statistical Regularization.* For more robustness, we use a statistically motivated regularization term that punishes shape and expression coefficients that deviate too much from the average:

$$E_{\text{regularizer}}(\alpha, \delta) = \sum_i \left\| \frac{\alpha_i}{\sigma_{i,\text{shape}}} \right\|_2^2 + \sum_i \left\| \frac{\delta_i}{\sigma_{i,\text{exp}}} \right\|_2^2 .$$

Here, $\sigma_{i,\text{shape}}$ and $\sigma_{i,\text{exp}}$ are the standard deviations of the corresponding shape and blendshape dimensions, respectively. The weighted sum of these three terms is minimized using the optimization method of Levenberg-Marquardt.

*Automatic Blendshape Transfer.* The established set of dense point-to-point correspondences allows us to build an expression basis for the person-specific actor rig by transferring the per-vertex blendshape displacements of the face model. The basis is only transferred inside a predefined face mask region, and if the correspondence lies within a threshold distance ($thres_{\text{transfer}} = 5\,\text{mm}$). We use a feathering operation to smoothly blend out the contribution of the transferred displacements close to the boundary of the mask. The feathering is predefined through an alpha mask on the face model. In addition, we transfer semantic information such as an eye region and a mouth region mask.

### 4.4 Parametric Body Model to Scan Fitting

In contrast to facial expressions, which are mostly linear, body motion is non-linear. To accommodate for this, we use a kinematic skeleton. We automatically rig the person-specific actor model by transferring the skinning weights and skeleton of a parametric body model. In our system, we use the *SMPL* [Loper et al. 2015] model. We perform a non-rigid model-based registration to the reconstructed 3D actor model, in a similar fashion as for the face. First, the required 6 sparse feature points are manually selected. These markers are used to initialize the shoulder position and the head position. We then solve for the 10 shape parameters and the joint angles of the kinematic chain of *SMPL*. After fitting, we establish a set of dense correspondences between the two models. Finally, we transfer the skinning weights, as well as the skeleton. We also use the correspondences to transfer body, neck and head region masks with corresponding feathering weights. Note, to ensure consistent skinning weights of neighboring vertices, we apply Gaussian smoothing (5 iterations of 1-ring filtering).

### 4.5 Tracking Refinement

To improve our results, we refine the per frame tracking information of the depth sequence based on our final parametric actor rig. To this end, we use the segmentation of the scan (head and body) and re-track the calibration sequence independently for both areas. This step compensates for miss-alignments in the initial tracking due to slight non-rigid motions of the target during capture. The refined tracking information leads to an improved quality of the following video-based rendering step.

## 5 VIDEO-BASED RENDERING

To synthesize novel portrait videos of the target actor, we apply video-based rendering with image data from the input video sequences and the tracked actor model as geometric proxy. With video-based rendering it is possible to generate photo-realistic novel views, in particular, we can correctly synthesize regions for which it is difficult to reconstruct geometry at a sufficiently high quality, i.e., hair. To achieve good results, we need good correspondence between the parametric 3D target actor rig and the video data captured in the calibration sequence, as they are obtained in our refined tracking stage (see Sec. 4.5). Based on these correspondences, we cross-project images from the input sequences to the projection of the deformed target actor model. We warp separately according to the torso and head motion, facial expression, and eye motion, and we take special care for the proper segmentation of fore- and background. An overview of our view-dependent image synthesis pipeline is shown in Fig. 4, and the single steps are described in the following sections.

### 5.1 Spatio-temporal Foreground Segmentation

First, we generate a foreground/background segmentation (Fig. 5) using a novel space-time graph cut approach (Fig. 6). We initialize the segmentation of the given image domain $\mathcal{I}$ by re-projecting the reconstructed and tracked proxy mesh to the calibration images to obtain an initial mask $\mathcal{M}$. Afterwards, we compute segmentation masks $\mathcal{F}, \mathcal{B}, \mathcal{U}_f$, and $\mathcal{U}_b$. $\mathcal{F}$ and $\mathcal{B}$ are confident foreground and background regions. Between them is an uncertainty region, with $\mathcal{U}_f$ being the probable foreground region, and $\mathcal{U}_b$ the probable background region.

The confident foreground region $\mathcal{F} = \mathcal{M} \ominus \mathcal{S}$ is computed by applying an erosion operator $\mathcal{M}$. The confident background $\mathcal{B} = \mathcal{I} \setminus (\mathcal{M} \oplus \mathcal{S})$ is the complement of the dilation of $\mathcal{M}$. In the remaining region of uncertainty, we perform background subtraction in HSV color space using a previously captured background image. If the pixel color differs from the background image more than a threshold, the pixel is assumed to be most likely a foreground pixel and assigned to $\mathcal{U}_f$, otherwise to $\mathcal{U}_b$. Finally, we remove outliers using a number of further erosion and dilation operations.

The resulting regions are used to initialize the GrabCut [Rother et al. 2004] segmentation algorithm[3]. Performing the segmentation per frame can result in temporally incoherent segmentation. Thus, we apply GrabCut to the entire 3D space-time volume of the calibration sequence. We do so by executing the approach independently on all $x$-, $y$-, and $t$-slices. The resulting foreground masks are combined in a consolidation step to generate the final foreground alpha mask (see Fig. 6).

### 5.2 Image Warping

Using the color data observed during the scanning process, we propose a view-dependent compositing strategy, see also Fig. 4. Based on the skinning weights, the body is clustered into body parts, which are textured independently. For each body part, we first retrieve the color frame of the calibration sequence that best matches its current modified orientation. We then initialize the per-view warp fields exploiting the morphed 3D geometry and cross-projection. To this end, we back-project the model into the retrieved frame using the tracking information. Then, we compute a warp field, i.e., a 2D displacement field in image space. The warp field maps from the re-projection in the retrieved image and the projection of the current model into screen space. Using a Laplacian image pyramid, we extend the warp field to the complete image domain. Finally, we use the extended warp field as described above and apply it to the retrieved image frames. Thus, we ensure that we also re-synthesize regions that are not directly covered by the proxy mesh, e.g., hair strands, and that we do not render parts of the mesh where actually the background is visible. The final per-region warps are blended based on a feathering operation using the body, neck, and head masks. Note, our image-based warping technique preserves the details from the calibration sequence since we select the texture based on the pose of the corresponding body part. This selection can be seen as a heuristic of finding the texture with minimal required warping to produce the output frame. Thus, detailed images with hair strands can be synthesized.

---

[2]http://www.cs.umd.edu/~mount/ANN/

[3]https://opencv.org

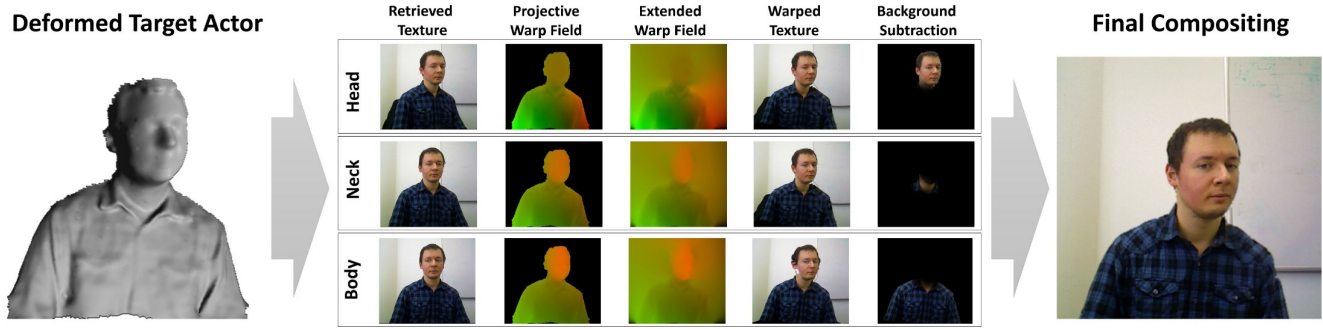

Fig. 4. Overview of the view-dependent image synthesis. Starting with a depth image of our target actor (left), we search for the closest frames in the input sequence, independently for the current head, neck, and body positions. For each such frame, a warp field is computed, and the frames are warped to the correct position. The warped images are then combined after a background subtraction and composited with the background to achieve a photo-realistic re-rendering. The shown uv displacements are color coded in the red and green color channel.

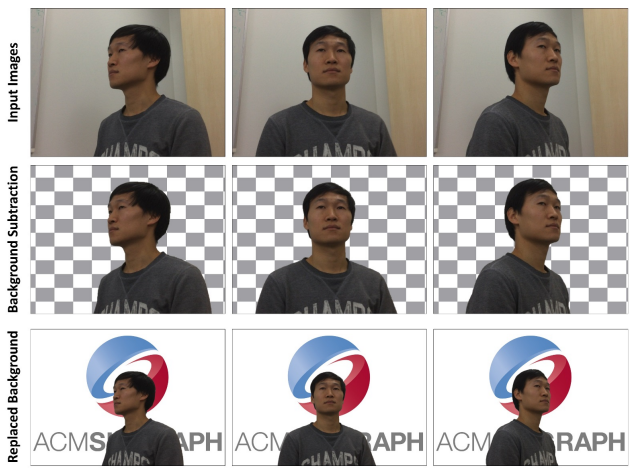

Fig. 5. Our temporal background subtraction: the top row shows the input color images and the middle row the extracted foreground layer using our space-time graph cut segmentation approach. The bottom row shows a background replacement example.

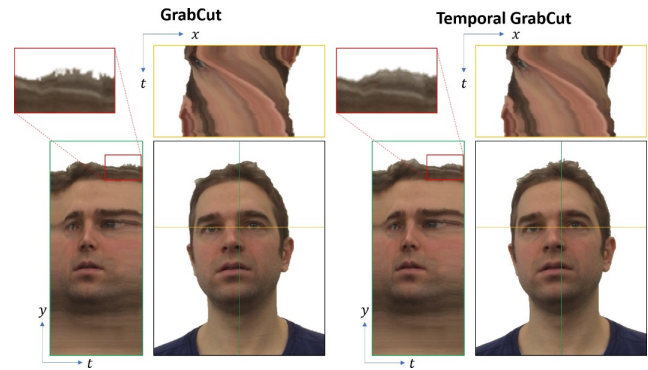

Fig. 6. Temporal GrabCut. On the left we show the output of the original GrabCut approach and on the right our temporal modified GrabCut. Our approach combines the segmentation results along the $xt$, $yt$ and $xy$ planes. The results on the left show the foreground masks retrieved from the $xy$ GrabCuts. Our extension of GrabCut to the temporal domain reduces flickering artifacts, thus, the foreground segmentations in the $xt$ and $yt$ planes are smoother.

## 6 REAL-TIME REENACTMENT

Our approach enables real-time reenactment of the head and torso in portrait videos. This requires real-time tracking of the source actor and an efficient technique to transfer the deformations from the source to the target. To this end, we apply our video-based rendering approach to re-render the modified target actor in a photo-realistic fashion. In the following, we detail our real-time upper body and face tracking approach, and describe the deformation transfer. In order to ensure real-time reenactment on a single consumer level computer, all components are required to run in a relatively short time span.

## 6.1 Source Tracking

We track the source actor using a monocular stream from a commodity RGB-D sensor (see Fig. 1). In our examples, we use either an Asus Xtion RGB-D sensor or a StructureIO sensor[4]. Our default option is the StructureIO sensor, which we set up for real-time streaming over WiFi in a similar configuration as Dai et al. [2017]. The StructureIO sensor uses the RGB camera of the iPad, allowing us to record the RGB stream at higher resolution ($1296 \times 968$) compared to the $640 \times 480$ resolution of the Asus Xtion. However, the WiFi streaming comes also with a latency of a few frames which is noticeable in the live sequences in the accompanying video, and the overall framerate is typically $20Hz$ due to the limited bandwidth.

The tracking of the source actor consists of two major parts, the face tracking and the upper body tracking as can be seen in Fig. 7.

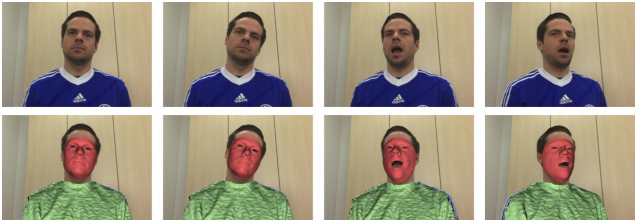

Fig. 7. Source actor tracking: Top: example input sequence of a source actor. Bottom: corresponding tracking results as overlay. The fitted face model is shown in red and the proxy mesh for tracking the upper body in green.

*6.1.1 Facial Performance Capture.* Facial performance capture is based on an analysis-by-synthesis approach [Thies et al. 2015] that fits the multi-linear face model that is also used for automatic rigging. We jointly optimize for the model parameters (shape, albedo, expression), rigid head pose, and illumination (first three bands of spherical harmonics) that best reproduce the input frame. The energy function is composed of a sparse landmark term that measures the distance of the model to detected 2D features (computed by the TVS marker tracker), a dense photometric appearance term that measures the color differences in RGB space, and a dense geometry term that considers point-to-point and point-to-plane distances from the model to the depth observations. For real-time performance, the resulting optimization problem is solved using a data-parallel Gauss-Newton solver. For more details on dense facial performance capture, we refer to Thies et al. [2015; 2016].

*6.1.2 Upper Body Tracking.* In order to track the upper body of the source actor within the limited computational time budget, we first compute a coarse mesh of the upper body. To achieve this mesh, we average a couple of depth frames that show the frontal facing source actor (about 20 frames). We use the tracking information of the face to determine the region of interest in this averaged depth map. That is, we segment the foreground from the background and use the region below the neck. We then extract the proxy mesh by applying a connected component analysis on the depth map. We track the rigid pose of the upper body with a model-to-frame ICP that uses dense projective correspondence association [Rusinkiewicz and Levoy 2001] and a point-to-plane distance measure.

---

[4]https://structure.io/

*6.1.3 Eye Gaze Tracking.* To estimate the eye gaze of the source actor, we use the TVS landmark tracker that detects the pupils and eye lid closure events. The 2D location of the pupils ($P_0, P_1 \in \mathbb{R}^2$, left and right pupil) are used to approximate the gaze of the person relative to the face model. We estimate the yaw angle of each eye by computing the relative position of pupil between the left ($C_{0,l}$) and right eye corner ($C_{0,r}$):

$$yaw_0 = \frac{||P_0 - C_{0,l}||_2}{||P_0 - C_{0,l}||_2 + ||P_0 - C_{0,r}||_2} \cdot 90° - 45° \ .$$

The pitch angle is computed in a similar fashion. We ignore squinting and vergence, and average the yaw and pitch angle of the left and right eye for higher stability. Finally, we map the yaw and pitch angle to a discrete gaze class that is defined by the eye calibration pattern, which was used to train the eye-synthesis for the target actor. If eye closing is detected, we overwrite the gaze class with the sampled closed eye class. This eye class can then be used to retrieve the correctly matching eye texture of the target rig.

## 6.2 Expression, Pose, and Gaze Transfer

Since the face model of the source actor uses the same blendshape basis as the target rig, we can directly copy the expression parameters. In addition, we apply the relative body deformations of the head, neck and torso to the corresponding joints of the kinematic skeleton of the target rig. These relative body deformations are computed via inverse kinematics using the tracked face and the tracked torso of the source actor. Since the rigid pose of the source and target actor is the same after applying the skeleton deformations, we can copy the mouth interior from the source to the target. In order to compensate for color and illumination differences, we use Poisson image editing [Pérez et al. 2003] with gradient mixing. We use predefined masks on the face template to determine the regions that must be copied and the areas where gradient mixing is applied (between the source image content and the synthesized target image). Using the eye class index estimated by our gaze tracker, we select the corresponding eye texture from the calibration sequence and insert the eye texture, again using Poisson image blending. To produce temporally smooth transitions between eye classes, we blend between the eye texture of the current and preceding frame. Fig. 8 shows the used textures and the extent of the eye and mouth blending masks that were applied to generate our reenactment results.

## 7 RESULTS

In this section, we test and evaluate our approach and compare to state-of-the-art image and video reenactment techniques. All following experiments have been performed on a single desktop computer with an Nvidia GTX1080 Ti and a 4.2GHz Intel Core i7-7700K processor.

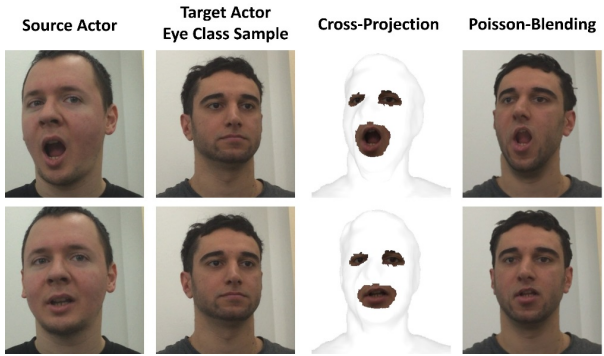

Fig. 8. Final compositing of the eye and mouth region; from left to right: driving frame of the source actor (used for mouth transfer), target actor eye class sample that corresponds to the estimated gaze direction of the source actor, cross-projection of the mouth and the eyes to the deformed target actor mesh, and the final composite based on Poisson image blending.

Table 1. Breakdown of the timings of the steps of our reenactment pipeline: dense face tracking (DenseFT), dense body tracking (DenseBT), deformation transfer (DT), morphing of the target actor mesh and image-based video synthesis (Synth), and cross projection and blending of the eyes and the mouth region. The first row shows timings for $640x480$ resolution (Asus Xtion) and the second row the timings for $1296x968$ (StructureIO).

|          | DenseFT  | DenseBT | DT      | Synth    | CB       | FPS    |
|----------|----------|---------|---------|----------|----------|--------|
| Avg.     | 10.91 ms | 1.34 ms | 1.13 ms | 4.41 ms  | 3.25 ms  | **47.5Hz** |
| Std.Dev. | 0.43     | 0.14    | 0.04    | 0.20     | 0.09     |        |
| Avg.     | 13.49 ms | 4.31 ms | 1.17 ms | 14.11 ms | 10.81 ms | **22.8Hz** |
| Std.Dev. | 0.43     | 0.21    | 0.09    | 0.31     | 0.21     |        |

Fig. 9 shows results from our live setup using the StructureIO sensor; please also see the accompanying video for live footage. As the results show, our approach generates high-quality reenactments of portrait videos, including the transfer of head pose, torso movement, facial expression, and eye gaze, for a large variety of source and target actors. The entire pipeline, from source actor tracking to video-based rendering, runs at real-time rates, and is thus applicable to interactive scenarios such as teleconferencing systems. A breakdown of the timings is shown in Tab. 1.

In the following, we further evaluate the quality of the synthesized video output and compare to recent state-of-the-art reenactment systems. Comparisons are also shown in the accompanying video.

*Evaluation of Video-based Rendering.* To evaluate the quality improvement due to our video-based rendering approach, we compare it with the direct rendering of the colored mesh obtained from the 3D reconstruction; see Fig. 10. Both scenarios use the same coarse geometry proxy that has been reconstructed using VoxelHashing [Nießner et al. 2013]. As can be seen, the video-based rendering approach leads to drastically higher quality compared to simple voxel-based colors. Since the proxy geometry can be incomplete, holes become visible in the baseline approach, e.g., around the ears and in the hair region. In our video-based rendering approach, these regions are filled in by our view- and pose-dependent rendering strategy using the extended warp field, producing complete and highly-realistic video output. Since the actor was scanned with closed mouth, opening of the mouth leads to severe artifacts in the baseline approach, while our mouth transfer strategy enables a plausible synthesis of the mouth region. Finally, note how the hair, including its silhouette is well reproduced.

*Evaluation of Eye Reenactment.* We compare our eye gaze reenactment strategy to the deep learning-based DeepWarp [Ganin et al. 2016] approach, which only allows for gaze editing. As Fig. 11 shows, we obtain results of similar quality if only gaze is redirected. Note, in contrast to our method, DeepWarp is not person specific, i.e., to re-synthesize realistically looking eyes, we need a calibration sequence.

*Photometric Error in Self Reenactment.* To evaluate the quality of our entire reenactment pipeline, we conducted a self-reenactment comparison. We first build a person-specific rig of a particular actor and then re-synthesize a sequence of the same actor. In this scenario, we can consider the source video as ground truth, and compare it with our synthesized result. Three frames of the comparison are shown in Fig. 12. The first image shows the reference pose, so this frame contains no error due to motion. Thus, the error of the first frame (0.04 $\ell_2$ distance in RGB color space) shows the error of our rerendering, and thus can be seen as baseline for the other frames. The average color difference error of the following frames is 0.0528, which is very close to this baseline. We assume that most of the additional error is due to the rigid misalignment of the head, which stems from the low-dimensional kinematic model. Please note that while the synthesized images do not match the ground truth perfectly, the visual quality of the results is nonetheless close to photo-real, and head and body pose are plausible.

*Comparison to* Face2Face. A comparison to Face2Face [Thies et al. 2016] is shown in Fig. 13. Face2Face only reenacts facial expression, and does not adapt head movement or eye gaze. Hence, the video flow of Face2Face often seems out-of-place, since the timing of all motions do not align, as noted by Suwajanakorn et al. [2017]. The effect is particularly visible in live videos, and it severely restricts the applicability to teleconferencing settings. Our approach achieves comparable quality of single frames, and generates more believable reenactment results by jointly re-targeting the rigid head pose, torso motion, facial expression, and eye gaze direction. Note that our technique copies the mouth from the source actor to the final output. Thus, the identity of the target person is slightly changed. Since Face2Face uses a database of mouth interiors of the target actor, the identity is unchanged. While it is straightforward to incorporate the mouth retrieval technique presented in Face2Face, we decided against it, because it drastically increases the length of the calibration phase and usability (since only mouth interiors that have been seen in the calibration sequence can be reproduced; note that also expressions with different rigid poses of the head would have to be captured in such a calibration).

*Comparison to* Bringing Portraits to Life. We also compare our method with Bringing Portraits to Life, an off-line image reenactment approach [Averbuch-Elor et al. 2017], which creates convincing reactive profile videos by transferring expressions and slight head motions of a driving sequence to a target image. It only requires a single image of the target actor as input, but does not provide any control over the torso motion and gaze direction. Fig. 14 shows results of the comparison. We achieve similar quality in general, but BRINGING PORTRAITS TO LIFE struggles for larger head pose changes. In comparison, our approach enables free head-pose changes, and provides control over the torso motion, facial expression, and gaze direction. Since our method runs at real-time rates, our approach can also be applied to live applications, such as teleconferencing.

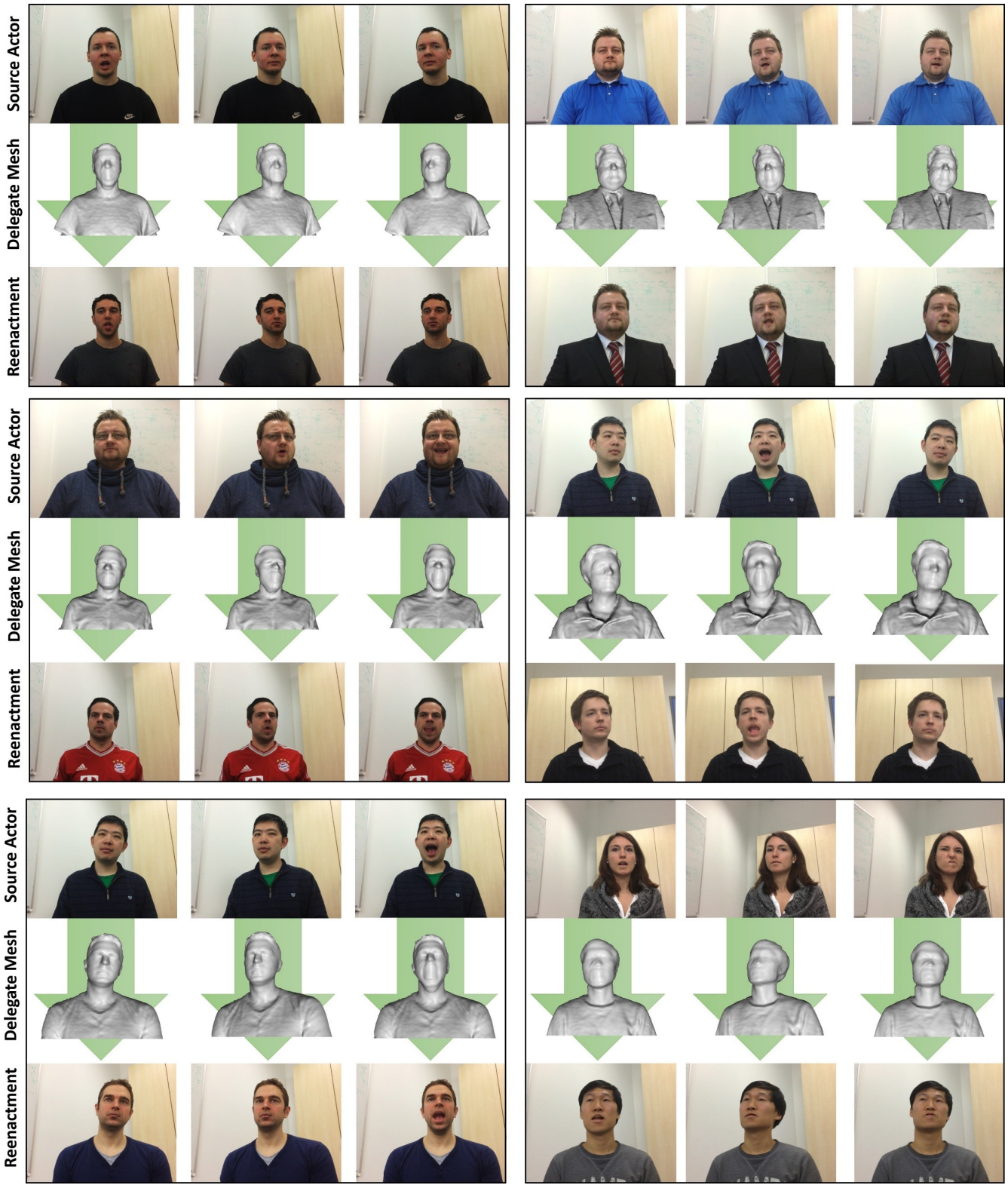

Fig. 9. Real-time portrait video reenactment results of our system for a variety of source and target actors. The source actor drives the head motion, torso movement, facial expression, and the eye gaze of the target actor in real time.

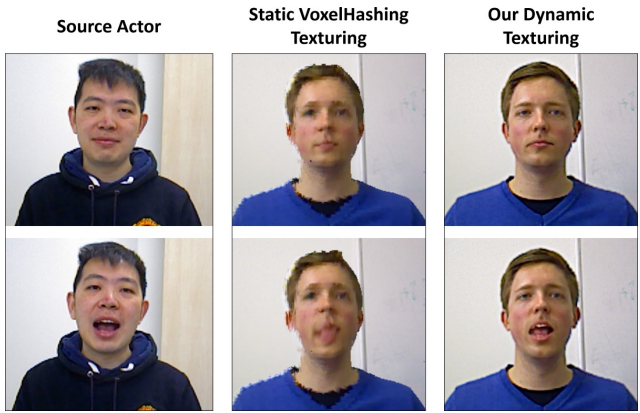

Fig. 10. Evaluation of video-based rendering: we compare our video-based rendering (right) and a simple colored-mesh actor proxy (middle). Both scenarios use the same coarse geometric proxy. Our video-based rendering approach leads to drastically higher realism in all regions and produces photo-realistic video output, while the colored-mesh lacks this fidelity.

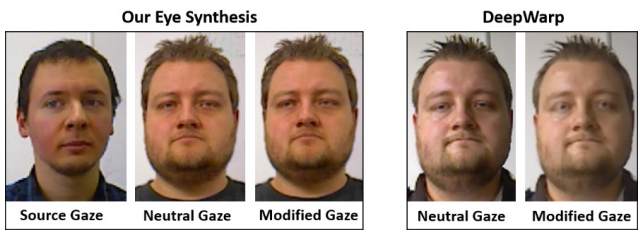

Fig. 11. Gaze redirection comparison: we compare our eye reenactment strategy (left) to the DeepWarp [Ganin et al. 2016] gaze redirection approach (right). Note that DeepWarp merely modifies gaze direction, but does not perform a full reenactment of portrait videos.

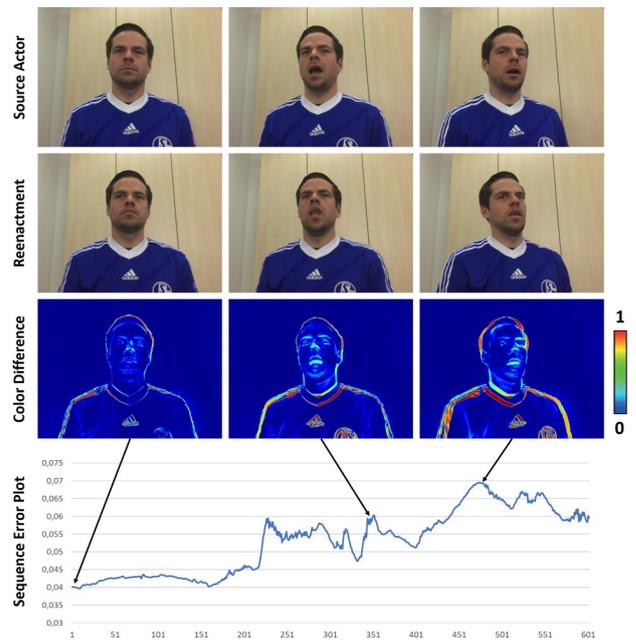

Fig. 12. Self-Reenactment Evaluation: the first column of the images shows the reference pose of the source and target actor; all following deformations are applied relative to this pose. For this experiment, we rigidly align the reference target actor body to the reference frame of the source actor in order to be able to compare the outputs. We compare the result to the source image using a per-pixel color difference measure. The other two columns show representative results of the test sequence with expression and pose changes. In the bottom row, the color difference plot of the complete test sequence is depicted. The mean $\ell_2$ color difference over the whole test sequence is 0.0528 measured in RGB color space ([0, 1]).

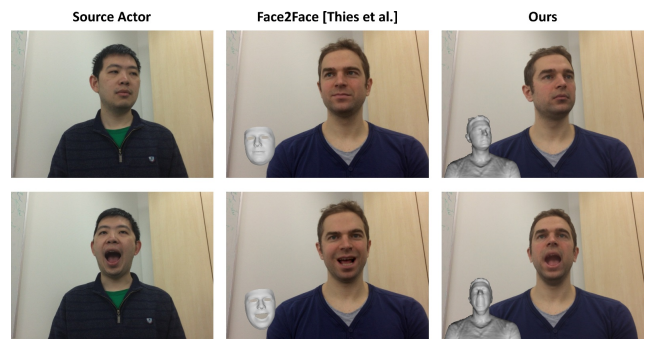

Fig. 13. Comparison to FACE2FACE [Thies et al. 2016]; from left to right: source actor, the reenactment result of FACE2FACE, and our result. In gray, we show the underlying geometry used to generate the output images.

*Comparison to* AVATAR DIGITIZATION. In Fig. 15, we also compare to the AVATAR DIGITIZATION approach of Hu et al. [2017]. From a single image, this approach generates an avatar, that can be animated and used for instance as a game character. However, the approach (as well as comparable avatar digitization approaches [Ichim et al. 2015]) generate stylized avatars that are appropriate as game-quality characters and that can be used in gaming and social VR applications. In contrast, we aim to synthesize unseen video footage of the target actor at photo-realistic quality as shown in Fig. 9.

## 8 LIMITATIONS

We have demonstrated robust source-to-target reenactment of complete human portrait videos at real-time rates. Still, a few limitations remain, and we hope that these are tackled in follow-up work. One drawback of our approach is the requirement of a short scanning sequence based on an RGB-D camera. While RGB-D sensors are already widespread, the ultimate goal would be to built the video-based target rig based on an unconstrained monocular video of the target actor, without a predefined calibration procedure. In addition, scene illumination is currently not estimated, and therefore illumination changes in reenacted videos cannot be simulated. We also do not track and transfer fine scale details such as wrinkles since they are not represented by the used multi-linear face model (see Fig. 16). While Cao et al. [Cao et al. 2015] demonstrate tracking of fine scale details, it has not be shown how to transfer these wrinkles to another person. This is an open question that can be tackled in the future. Under extreme pose changes, or difficult motion of hair (see Fig. 18), the reenacted results may exhibit artifacts as neither the model nor the video-based texturing may be able to fully represent the new view-dependent appearance. In Fig. 17 we show failure cases that stem from extreme head rotations and occlusions in the input stream of the source actor. Note that the proposed technique has the same limitations as other state-of-the-art reenactment methods like Face2Face [Thies et al. 2016]. In particular, the used analysis-by-synthesis approach to track the face uses the parameters of the previous frame as an initial guess, thus, fast head motions require high frame rates of the input camera otherwise the tracking is disturbed by the motion (for more details on the limitations of the face tracking we refer to the publications [Thies et al. 2015, 2016]). Our approach is also limited to the upper body. We do not track the motions of the arms and hands, and are not able to re-synthesize such motions for the target actor. Ideally, one would want to control the whole body; here, we see our project as a stepping stone towards this direction, which we believe will lead to exciting follow up work. We do believe that the combination of a coarse deformation proxy with view-dependent textures will generalize to larger parts of the body, if they can be robustly tracked.

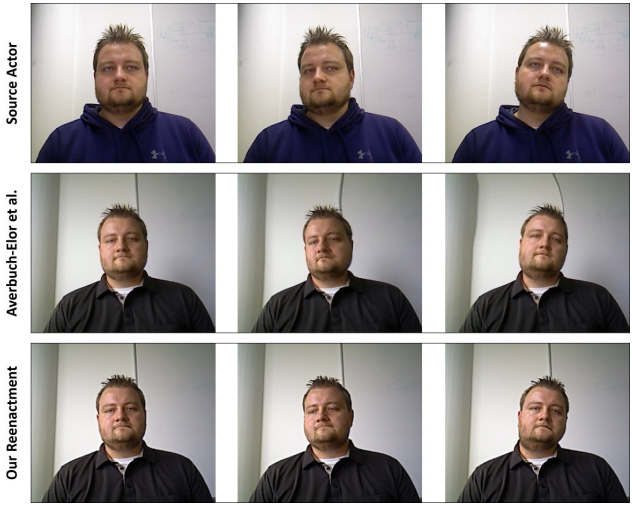

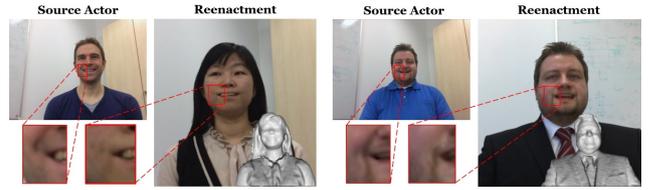

Fig. 16. Limitation: Fine scale detail such as wrinkles are not transferred. The close-ups show the difference between the input and the output.

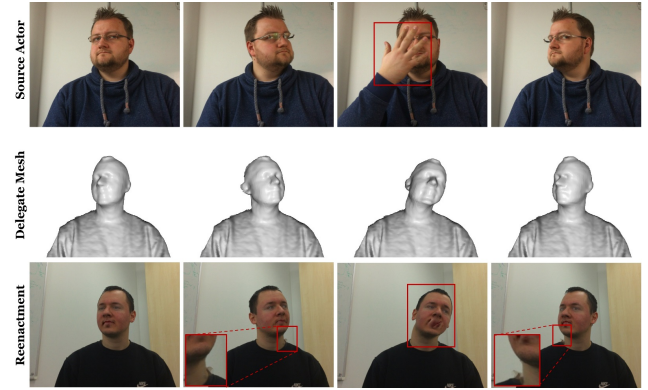

Fig. 17. Limitation: Strong head rotations or occlusions in the input stream of the source actor lead to distortions in the reenactment result.

Fig. 14. Comparison to Bringing Portraits to Life [Averbuch-Elor et al. 2017]: Our approach generalizes better to larger changes in head and body pose than the image-warping based approach of Averbuch-Elor et al. [2017]. In addition, our methods enables the joint modification and control of the torso motion and gaze direction. Note that while their approach runs offline, ours allows control the entire portrait video at real-time frame rates, allowing application to live teleconferencing.

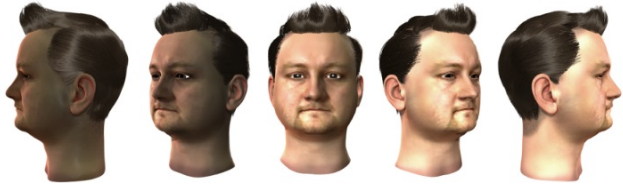

Fig. 15. Avatar Digitization reconstructs stylized game-quality characters from a single image. In this example, the avatar was generated from the first image of the second row in Fig. 14.

## 9 CONCLUSION

We introduced HeadOn, an interactive reenactment system for human portrait videos. We capture facial expressions, eye gaze, rigid head pose, and motions of the upper body of a source actor, and transfer them to a target actor in real time. By transferring all relevant motions from a human portrait video, we achieve believable and plausible reenactments, which opens up the avenue for many important applications such as movie editing and video conferencing. In particular, we show examples where a person is able to control portraits of another person or to perform self-reenactment to easily switch clothing in a live video stream. However, more fundamentally, we believe that our method is a stepping stone towards a much broader avenue in movie editing. We believe that the idea of coarse geometric proxies can be applied to more sophisticated environments, such as complex movie settings, and ultimately transform current video processing pipelines. In this spirit, we are convinced and hopeful to see many more future research works in this exciting area.

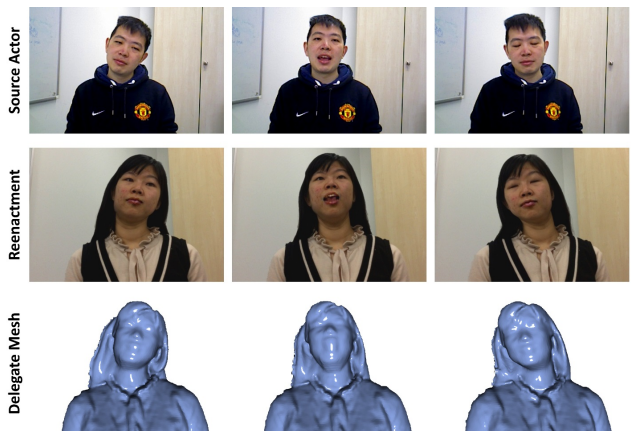

Fig. 18. Limitation: Hair is statically attached to the skeleton structure of the delegate mesh.

## ACKNOWLEDGMENTS

We thank Angela Dai for the video voice over and all actors for participating in this project. Thanks to Averbuch-Elor et al. and Hu et al. for the comparisons. The facial landmark tracker was kindly provided by TrueVisionSolution. This work was supported by the ERC Starting Grant CapReal (335545), the Max Planck Center for Visual Computing and Communication (MPC-VCC), a TUM-IAS Rudolf Mößbauer Fellowship, and a Google Faculty Award.